\documentclass[sigconf,nonacm]{acmart}

\AtBeginDocument{%
  }

\setcopyright{acmlicensed}
\copyrightyear{2018}
\acmYear{2018}
\acmDOI{XXXXXXX.XXXXXXX}
\acmConference[Conference acronym 'XX]{Make sure to enter the correct
  conference title from your rights confirmation email}{June 03--05,
  2018}{Woodstock, NY}
\acmISBN{978-1-4503-XXXX-X/2018/06}

\usepackage[greek,english]{babel}
\usepackage{textgreek}  
\usepackage{alphabeta}  
\usepackage{tabularx}

\begin{document}

\title{A Greek Government Decisions Dataset for Public-Sector Analysis and Insight}


\author{
Giorgos Antoniou \quad
Giorgos Filandrianos \quad
Aggelos Vlachos \quad
Giorgos Stamou
}
\affiliation{%
  \institution{National Technical University of Athens, Greece}
  \country{}}
\email{georgiosantoniou@mail.ntua.gr, geofila@ails.ece.ntua.gr}
\email{aavlachos@cslab.ece.ntua.gr, }

\author{Lampros Kollimenos}
\affiliation{%
  \institution{University of Thessaly, Greece}
  \country{}}
\email{lkollimenos2001@gmail.com}

\author{Konstantinos Skianis}
\affiliation{%
  \institution{University of Ioannina, Greece}
  \country{}
}
\email{kskianis@cse.uoi.gr}

\author{Michalis Vazirgiannis}
\affiliation{%
  \institution{Ecole Polytechnique, France}
  \country{}
  }
\email{mvazirg@lix.polytechnique.fr}


\renewcommand{\shortauthors}{Antoniou et al.}

\begin{abstract}
We introduce an open, machine-readable corpus of Greek government decisions sourced from the national transparency platform \textit{Diavgeia}\footnote{\url{https://diavgeia.gov.gr/}}.
The resource comprises 1 million decisions, featuring and high-quality raw text extracted from PDFs.
It is released with raw extracted text in Markdown format, alongside a fully reproducible extraction pipeline.

Beyond the core dataset, we conduct qualitative analyses to explore boilerplate patterns and design a retrieval-augmented generation (RAG) task by formulating a set of representative questions, creating high-quality answers, and evaluating a baseline RAG system on its ability to retrieve and reason over public decisions. 
This evaluation demonstrates the potential of large-scale public-sector corpora to support advanced information access and transparency through structured retrieval and reasoning over governmental documents, and highlights how such a RAG pipeline could simulate a chat-based assistant capable of interactively answering questions about public decisions.

Due to its scale, quality, and domain coverage, the corpus can also serve as high-value pre-training or fine-tuning material for new Language Models (LMs) and Large Language Models (LLMs) respectively, including specialized models for legal and governmental domains, and as a foundation for novel approaches in domain adaptation, knowledge-grounded generation, and explainable AI. 
Finally, we discuss limitations, outline future directions, and make both the data and the code accessible \footnote{\url{https://github.com/y3nk0/diavgeia}}.
\end{abstract}



\received{20 February 2007}
\received[revised]{12 March 2009}
\received[accepted]{5 June 2009}

\maketitle

\section{Introduction}

Transparency, accountability, and data-driven supervision of public institutions are made possible by open government data, which has emerged as a key component of contemporary democratic governance.  Governments can lessen information asymmetries, empower civil society, and encourage evidence-based policy evaluation by making administrative records accessible and machine-readable \cite{zuiderwijk2014opendata, bertot2010transparency}. 
 Large-scale open-data programs throughout Europe and beyond have shown that open disclosure of official choices may boost anti-corruption efforts, increase public service quality, and build confidence \cite{edp2020maturity}. 
 One notable example of such institutional commitments is Greece's national transparency platform \textit{Diavgeia}, which requires the disclosure of all administrative decisions made by public agencies, municipalities, and ministries.

Documents and data provided by \textit{Diavgeia} differ from Greek law datasets because they capture the everyday operational decisions of the public administration, not the formal legal framework enacted by Parliament. 
Some well-known and widely used publicly available Greek law datasets and legislation repositories include: i) “Published laws from the National Printing Office\footnote{\url{https://data.gov.gr}}” includes metadata, issue numbers, etc.; ii) The official legislation search portal of Hellenic Parliament\footnote{\url{https://www.hellenicparliament.gr/en/nomothetiko-ergo/anazitisi-nomothetikou-ergou}} for finding laws, bills, and enacted legislation; iii) the National Printing Office of Greece (Official Gazette / FEK)\footnote{\url{https://search.et.gr/en/search-legislation}} search interface — for browsing the full official Government Gazette issues; iv)A consolidated-law dataset on Hugging Face: “Permanent Greek Legislation Code - Raptarchis\footnote{\url{https://huggingface.co/datasets/AI-team-UoA/greek_legal_code}}”, a dataset consisting of approximately 47k legal resources of Greek legislation from 1834 to 2015 and is suitable for research/machine learning (ML) tasks.
Laws and presidential decrees are structured, hierarchical, stable legal texts produced through a formal legislative process, while \textit{Diavgeia} contains ministerial acts, procurement decisions, budget allocations, hiring approvals, board resolutions, and thousands of other administrative actions issued continuously by agencies across the state. 
As a result, \textit{Diavgeia} is massive, dynamic, heterogeneous, and closer to administrative reality, whereas laws represent the normative legal order. 
This makes \textit{Diavgeia} uniquely valuable for transparency, auditing, and empirical governance analysis, offering a granular, real-time view of how the public sector operates—far beyond what legal corpora alone can provide.

However, the existence of open data alone does not guarantee accessibility or meaningful public use. 
Government documents are often dispersed across formats such as PDFs and JSONs, inconsistently structured, and difficult for non-experts to interpret. 
Recent advances in natural language processing (NLP), information extraction, and retrieval systems provide new opportunities to transform raw administrative text into searchable, understandable, and actionable knowledge \cite{chalkidis2019large, chalkidis2022lexglue}. 
Systems that leverage domain-adapted language models, question answering, or retrieval-augmented generation can substantially improve how citizens navigate public decisions, how journalists investigate patterns of spending or policy shifts, and how researchers analyze institutional behavior \cite{lewis2020rag}. 
When deployed responsibly, such technologies offer a path toward more inclusive digital governance and stronger democratic participation.

In this work, we introduce the first large-scale, curated, and machine-readable corpus of Greek government decisions sourced from the \textit{Diavgeia} platform. 
Our main contributions are:

\begin{itemize}
    \item \textbf{A large-scale open dataset:} 1 million government decisions with normalized metadata, canonical identifiers, and high-quality Markdown text extracted from PDFs/JSONs, released in interoperable formats.
    \item \textbf{A fully reproducible extraction pipeline:} End-to-end tooling for harvesting, cleaning, normalizing, and structuring \textit{Diavgeia} documents at scale.
    \item \textbf{Boilerplate analysis:} Empirical exploration of administrative patterns, decision types, institutional behaviors, and textual characteristics across agencies.
    \item \textbf{A retrieval-augmented generation (RAG) task:} A benchmark comprising representative questions, curated answers, and baseline RAG performance over the dataset.
\end{itemize}

This combination of open data infrastructure, analytical insights, and practical tools establishes a foundation for future work in legal NLP, digital governance, and democratic technology.

\section{Related Work}

Open government data has long been recognized as a foundation for transparency, accountability, and civic engagement. 
Early work established that making governmental documents publicly accessible enables independent auditing, reduces information asymmetries, and fosters trust in democratic institutions \cite{zuiderwijk2014opendata, bertot2010transparency}. 
Large-scale initiatives such as the European Data Portal have demonstrated the importance of standardisation and interoperability in realising the full societal value of public data \cite{edp2020maturity}. 
In the Greek context, the \textit{Diavgeia} transparency portal represents one of the most ambitious national-level efforts, mandating the publication of all administrative decisions. 
Yet despite its legal significance and substantial scale, technical exploitation of its content has historically been limited.

Early efforts to exploit \textit{Diavgeia} leaned primarily on semantic web tooling rather than full NLP.  
The civic-tech platform \textit{PublicSpending.gr} semantically processed daily spending feeds to generate citizen-facing visualisations, although it focused mainly on descriptive analytics rather than deeper modelling \citep{vafopoulos2012publicspending}.  
More recently, the \textit{d2kg} project proposed an OWL ontology that transforms \textit{Diavgeia} decisions into a linked-data graph, enabling SPARQL-based retrieval across acts and legal references, yet leaving textual patterns, administrative intent, and authorial practice largely unexplored \citep{serderidis2024d2kg}.  
Comparable open-government corpora abroad provide methodological blueprints that \textit{Diavgeia} research can adopt.  
The \textit{TheyBuyForYou} knowledge graph integrates European procurement notices to study anomalies in public spending \citep{soylu2021theybuyforyou}, while Germany’s NIF4OGGD converts federal open-data dumps into the NLP Interchange Format, improving cross-dataset search and text reuse detection \citep{sherif2014nif4oggd}.  
Rich legislative speech resources such as GerParCor Reloaded \citep{abrami2024german} and the ParlaMint Widened FOIA corpus \citep{viira2024parlamint} further illustrate how consistent schema design and preprocessing unlock large-scale diachronic political analyses.

Parallel to these open-data initiatives, the NLP community has made significant advances in processing legal and governmental documents. Recent work in legal NLP demonstrates the effectiveness of machine learning for multi-label classification, named-entity recognition, argument extraction, and domain-specific information retrieval \cite{chalkidis2019large}. 
The growing availability of legal benchmarks, such as LexGLUE \cite{chalkidis2022lexglue}, and the emergence of retrieval-augmented generation (RAG) systems tailored to complex knowledge-intensive tasks \cite{lewis2020rag} have created opportunities to build citizen-oriented tools that simplify legal language and support public understanding of institutional decision-making.

Although \textit{Diavgeia} represents a landmark for transparent governance, broader challenges identified in e-government studies remain relevant.  
Recent work highlights persistent issues around user experience, interface complexity, and data accessibility in general across public digital services \cite{beris2018modeling, stylianou2022doc2kg}.  
Many individuals struggle to navigate large document collections or locate relevant decisions, while users with disabilities or limited digital literacy face additional barriers.  
These usability constraints limit the platform's effectiveness in promoting transparency and enabling meaningful civic participation—underscoring the need for machine-readable resources, robust NLP pipelines, and citizen-facing retrieval or summarization tools.

\section{Dataset}

In this Section, we present the dataset, extraction techniques, final extracted data, and computed statistics.

\subsection{\textit{Diavgeia} API and Initial Data}
Currently \textit{Diavgeia} offers more than 71 million decisions.
Using the \textit{Diavgeia} OpenData API\footnote{\url{https://diavgeia.gov.gr/api/help}}, a user is allowed to download a number of decisions, respecting specific request limits.
The API provides a structured dataset comprising essential details of governmental decisions, including administrative metadata, decision dates, and legal references. 
This structured format ensures the data's reliability and accessibility, facilitating comprehensive analysis and insights into governmental operations.
Each decision comes with structured info in an JSON format, along with the PDF.
The main structure fields of a decision in the API is shown in Table \ref{tab:diavgeia-data} of the Appendix.
We downloaded 1 million decisions from the year 2021, with the total size of PDFs and metadata going over 1TB of storage space. 
It is obvious that the size of this dataset, make it very difficult to be handled, and thus transforming it to text is the best approach to manage it efficiently.
The size of the final extracted text files in compressed format is 4GBs, making it considerably more manageable than the original PDFs.
We note that the data include personal names as well as names of entities and organizations. Since these records are publicly available, we do not perform any removal of personally identifiable information.

\subsection{Text Extraction Comparative Analysis}
Since the extraction process is very important in datasets like these, we also present a text extraction comparative analysis of different approaches.
In order to further understand the text extraction process, we compare the performance of four methods: i) Tesseract OCR (CPU)\footnote{\url{https://github.com/tesseract-ocr/tesseract}}, ii) Paddle OCR (GPU)\footnote{\url{https://github.com/PaddlePaddle/PaddleOCR}}, iii) Direct PDF Extraction (PyMuPDF4LLM\footnote{\url{https://pymupdf.readthedocs.io/en/latest/pymupdf4llm/}}), and iv) an LLM-generated "Ground Truth" (GPT-5-nano), on a dataset of 200 Greek administrative PDF files.
The primary goal was to find the most accurate method for text extraction, given the known issues with broken encoding in the source PDFs, which makes direct extraction unreliable.
Due to the size of dataset, the approach we used PyMuPDF4LLM, mainly for two reasons: i) it was considerably faster compared to the others, and ii) the output is given in a Markdown format, which is considered optimal for training language models.
An extensive analysis of time performance is presented in Table \ref{tab:ocr_speed_comparison} of the Appendix.

\subsection{Corpus-level Statistics}

\begin{table}[ht]
\centering
\caption{Corpus Statistics for 1M \textit{Diavgeia} documents.}
\label{tab:diavgeia_corpus_stats}
\begin{tabular}{l r}
\hline
\textbf{Metric} & \textbf{Value} \\
\hline
Number of Documents & 1{,}061{,}438 \\
Total Tokens & 1{,}495{,}839{,}382 \\
Unique Documents Read & 1{,}061{,}438 \\
Average Tokens per Document & 5{,}455.62 \\
Median Tokens per Document & 2{,}806 \\
Standard Deviation (Tokens per Document) & 36{,}414.89 \\
Maximum Tokens in a Document & 19{,}169{,}263 \\
\hline
Total Characters & 2{,}005{,}743{,}579 \\
Average Characters per Document & 5{,}936.01 \\
Total Sentences & 38{,}740{,}934 \\
Average Sentences per Document & 36.50 \\
\hline
\end{tabular}
\end{table}

The corpus statistics were computed using a high-performance processing pipeline designed for large-scale text analytics. All documents were pre-extracted as plain text, and tokenization was performed with the Rust-optimized tiktoken\footnote{\url{https://github.com/openai/tiktoken}} library, which provides fast, LLM-compatible subword tokenization. 
To handle over one million files efficiently, we employed Python multiprocessing with batch dispatching across all CPU cores, enabling parallel reading and tokenization of documents. Additional metrics—such as character counts and sentence estimates—were computed using lightweight regular-expression processing to avoid the overhead of full NLP parsing. The combination of Rust-based tokenization and multi-core processing ensured that the entire \textit{Diavgeia} corpus could be analyzed efficiently at scale.

The \textit{Diavgeia} corpus comprises over 1 million documents totaling approximately 1.5 billion tokens, making it one of the largest publicly available Greek-language administrative datasets. The average document length is substantial ($\sim$5,456 tokens), with a median of 2,806 and a very high maximum of 19 million tokens, reflecting the presence of long legal annexes, procurement contracts, technical descriptions, and multi-page governmental reports. The high standard deviation (36,415 tokens) confirms the extreme heterogeneity of document sizes, ranging from minimal administrative notices to highly detailed financial or regulatory texts. The corpus also contains over 2 billion characters and nearly 39 million sentences, indicating rich linguistic variety and significant complexity in sentence structure.
All extracted statistics are shown in Table \ref{tab:diavgeia_corpus_stats}.

From an NLP perspective, these characteristics make \textit{Diavgeia} particularly valuable for training and evaluating models on long-context processing, financial reasoning, multi-document aggregation, legal-style summarization, and entity extraction. 
The scale of 1.5B tokens is more than sufficient for continued pretraining or domain adaptation of modern small-to-medium LLMs, significantly improving performance on Greek public-administration tasks. At the same time, the heterogeneity and vocabulary richness pose challenges for tokenization, retrieval, and hallucination control, making this corpus a strong benchmark for robustness, RAG evaluation, and long-form sequence modeling. 
In short, the \textit{Diavgeia} corpus is large and diverse enough to meaningfully support both applied NLP systems and research into domain specialization for LLMs.

\begin{table}[ht]
\centering
\caption{Top 5 organizations by number of \textit{Diavgeia} decisions}
\label{tab:top5_orgs}
\begin{tabular}{l r}
\hline
\textbf{Organization} & \textbf{\# Docs} \\
\hline
Aristotle University of Thessaloniki & 24{,}997 \\
National \& Kapodistrian University of Athens & 22{,}235 \\
National Technical Chamber & 17{,}412 \\
Unified Social Security Institution & 16{,}385 \\
Ministry of Culture and Sports & 15{,}383 \\
\hline
\end{tabular}
\end{table}

Table \ref{tab:top5_orgs} shows the top 5 organizations by number of \textit{Diavgeia} decisions.
These organizations are all large, central actors in the Greek public sector: two major universities (1st and 2nd), the national technical chamber (3rd), the unified social security institution (Εθνικός Φορέας Κοινωνικής Ασφάλισης in Greek) coming 4th, and the Ministry of Culture and Sports (Υπουργείο Πολιτισμού και Αθλητισμού) as 5th. 
They manage substantial budgets, infrastructure, and personnel, and are involved in continuous operational activity (procurement, project funding, appointments, committees, travel approvals, cultural events, maintenance, etc.). 
Because \textit{Diavgeia} mandates publication of a wide range of such administrative acts, high-volume institutions naturally generate a very large number of decisions, which explains why they dominate the corpus in terms of document count.

\section{Qualitative Analysis: Boilerplate Study}

\begin{figure}[h]
  \centering
  \includegraphics[width=\linewidth]{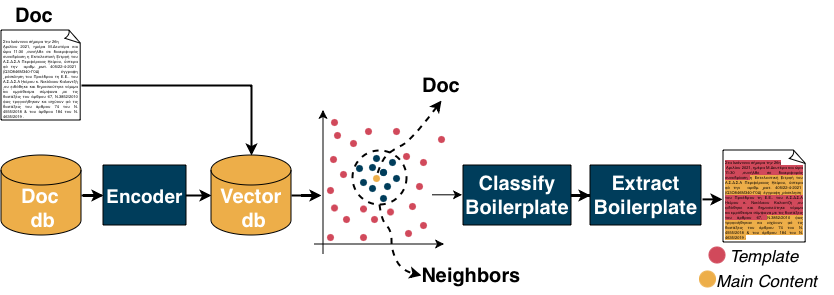}
  \caption{Our Boilerplate detection pipeline.}
  \label{fig:boiler}
\end{figure}

Boilerplate removal plays a critical role in transforming raw administrative documents into information that is more accessible and meaningful for end users. Government decisions published on \textit{Diavgeia} are often dense, technical, and formatted primarily for bureaucratic workflows rather than public readability, resulting in long documents where essential information is buried beneath recurring legal templates, procedural phrasing, and structural overhead. While boilerplate extraction has been explored in English \cite{leonhardt2020boilerplate, vogels2018web2text}, the space remains significantly underexplored in for Greek administrative text, where document patterns, legal phrasing, and structural conventions differ markedly. An effective boilerplate removal system could automatically identify and separate core decision content from recurring legal scaffolding, enabling users to focus on the substantive elements of each document without navigating repetitive language. In this work, we introduce a Greek-language boilerplate extraction approach using LLM-based modelling, aiming to reduce textual friction and improve the interpretability of public-sector documents at scale.

\subsection{Boilerplate Algorithm}
Our method is illustrated in Figure \ref{fig:boiler}. Each document is first stored in a structured document repository, from which its full textual stream is processed through an encoder to obtain fixed-length dense embeddings\footnote{Embeddings were generated using \href{https://huggingface.co/sentence-transformers/all-MiniLM-L6-v2}{all-MiniLM-L6-v2}.} . These representations are indexed into a vector database, enabling efficient k-NN retrieval across the corpus. At inference time, a query document is projected into the same embedding space and its nearest neighbours are obtained via cosine similarity. Semantically proximal documents are expected to share recurring administrative phrasing and structural patterns, allowing neighbourhood density to function as an implicit indicator of boilerplate presence.

Crucially, we intentionally avoid lexical similarity metrics such as keyword overlap, BM25 scoring \cite{chen2023bm25}, or edit-distance measures (e.g., Levenshtein), as these techniques overweight surface-form correspondence (see more in Section. Specifically, boilerplate templates can be short in proportion to the varying main body, which means lexical comparators tend to be dominated by document-specific content rather than structural repetition. As a result, vocabulary - based signals amplify noise instead of revealing template-consistency. Semantic embeddings, however, are better suited to the task: even when phrasing varies, documents derived from the same template frequently reference to similar public organisations, statutes, funding mechanisms or procedural entities. Thus, shared meaning rather than shared wording becomes the discriminative feature,  enabling template-based clustering to emerge more naturally in embedding space \ref{sec:boiler-prev-analysis}.). 
The final processing stage consists of two LLM-based modules. The first model performs boilerplate classification, estimating the likelihood that the source document (or a segment thereof) originates from a reusable administrative template. The second model conducts boilerplate extraction, returning a segmented form of the input text in which each span is labelled either as template boilerplate or main content. Unlike highlight-based visual presentation, this stage yields a structural decomposition of the document, explicitly disentangling invariant legal scaffolding from case-specific informational material. Importantly, segmentation is carried out dynamically on a per-document basis, acknowledging that template magnitude, phrasing and placement vary significantly across decisions.

\subsection{Evaluation}

To assess the effectiveness of our boilerplate extraction framework, we design an intrinsic evaluation procedure grounded in the following hypothesis. If two documents share the same boilerplate template and this template is correctly identified, then swapping the extracted main content between them should yield two documents structurally identical to the originals.

To obtain a reliable ground-truth basis for testing this hypothesis, we first conducted a manual annotation phase. Human annotators examined a subset of \textit{Diavgeia} decisions and grouped together documents that were judged to share the same underlying administrative template. From these human-validated template clusters, we constructed an evaluation sample consisting of 100 document pairs.

For each document, our algorithm then extracts the boilerplate skeleton and the corresponding main content. We subsequently perform cross-swapping of the content segments within each pair. Under correct boilerplate segmentation, the reconstructed document should closely approximate its counterpart, since both are expected to differ only in their variable, non-boilerplate sections.

To quantify reconstruction quality, we compute the normalized word-level Levenshtein distance between the reconstructed document and the reference target, treating edit similarity as a measure of extraction accuracy. Lower distance values indicate higher structural alignment, reflecting successful separation of template and content. In addition, we compute the Boilerplate Extraction Rate (BER), defined as the ratio of correctly recovered boilerplate tokens to the total tokens in the document. This allowing us to measure how completely each system identifies and isolates the underlying template. The reconstruction error (RE) and the BER using GPT-5-mini\footnote{gpt-5-mini-2025-08-07}  and GPT-5\footnote{gpt-5-2025-08-07} based extraction models are reported in Table~\ref{tab:reconstruction-error}.

\begin{table}[h]
\centering
\caption{Reconstruction error after boilerplate extraction and cross-content substitution.}
\label{tab:reconstruction-error}
\begin{tabular}{l|cc}
\toprule
\textbf{Model} & \textbf{RE ↓} & \textbf{BER (\%) ↑}  \\
\midrule
GPT-5-mini  & $0.0203_{\pm 0.0445}$ & $58.19_{\pm  29.57}$\\
GPT-5       & $0.0097_{\pm 0.0370}$ & $75.60_{\pm 24.71}$\\
\bottomrule
\end{tabular}
\end{table}

The results in Table~\ref{tab:reconstruction-error} show that both models are able to recover boilerplate templates with very high fidelity, as evidenced by the extremely low RE values. For GPT-5, an RE of 0.0097 at the word level implies that, for a document of 100 words, the reconstructed version differs on average in fewer than a single word-level edit (insertions, deletions or substitutions), i.e., under 1\% of the text is affected after template extraction and content swapping. Even GPT-5-mini, with an RE of 0.0203, remains within a small deviation band of roughly 2\% of the words. In conjunction with the high BER values observed, these results indicate not only minimal distortion during reconstruction, but also substantial boilerplate capture, meaning that the models recover a large share of template text and do so with very little structural loss.

Looking at boilerplate extraction performance overall, GPT-5 clearly outperforms GPT-5-mini, recovering on average 75.60\% of boilerplate content compared to 58.19\%, while simultaneously achieving lower RE. This indicates that GPT-5 not only preserves the underlying document structure almost perfectly, but also captures a substantially larger fraction of templated spans, resulting in a more complete and reliable separation between boilerplate and variable content.

\subsection{Boilerplate Prevalence Analysis}
\label{sec:boiler-prev-analysis}

\begin{figure}[!h]
    \centering
    \includegraphics[width=0.95\linewidth]{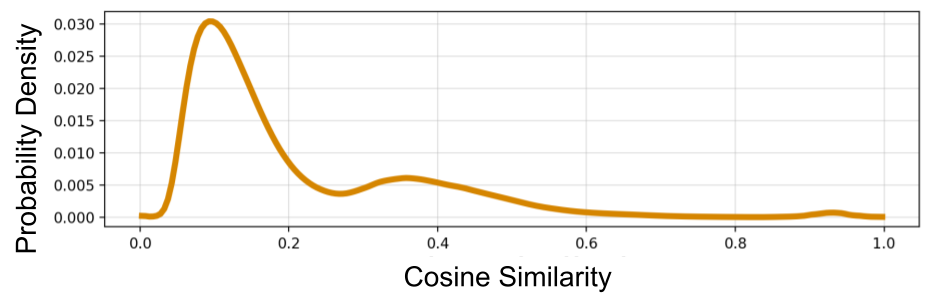}
    \caption{Pairwise cosine-distance distribution across document embeddings.}
    \label{fig:distance-distribution}
\end{figure}

To understand the structural characteristics of the corpus, we begin by estimating the degree to which boilerplate language appears within \textit{Diavgeia} documents. Since running full-scale extraction across the entire dataset is computationally expensive due to the LLM components of our pipeline, we instead adopt an alternative exploratory strategy that also offers methodological insight into the behaviour and assumptions of the algorithm outlined above. 
Figure \ref{fig:distance-distribution} illustrates the distribution of pairwise cosine distances between the document embeddings, providing an overview of the similarity relationships across the corpus. Distances were computed directly in embedding space, revealing how closely textual units are positioned relative to one another. The curve exhibits a sharp concentration of low-distance pairs, suggesting that individual documents tend to have many highly similar neighbours and comparatively fewer distant counterparts. Such behaviour is indicative of recurrent linguistic segments and duplicated structural patterns, where groups of documents share large overlapping spans of text. This clustering effect aligns with the presence of boilerplate material: if multiple documents are consistently situated near one another in semantic space, it is likely due to shared administrative templates, repeated procedural phrasing, or legally mandated formulations. 

\begin{figure}[!h]
    \centering
    \includegraphics[width=0.95\linewidth]{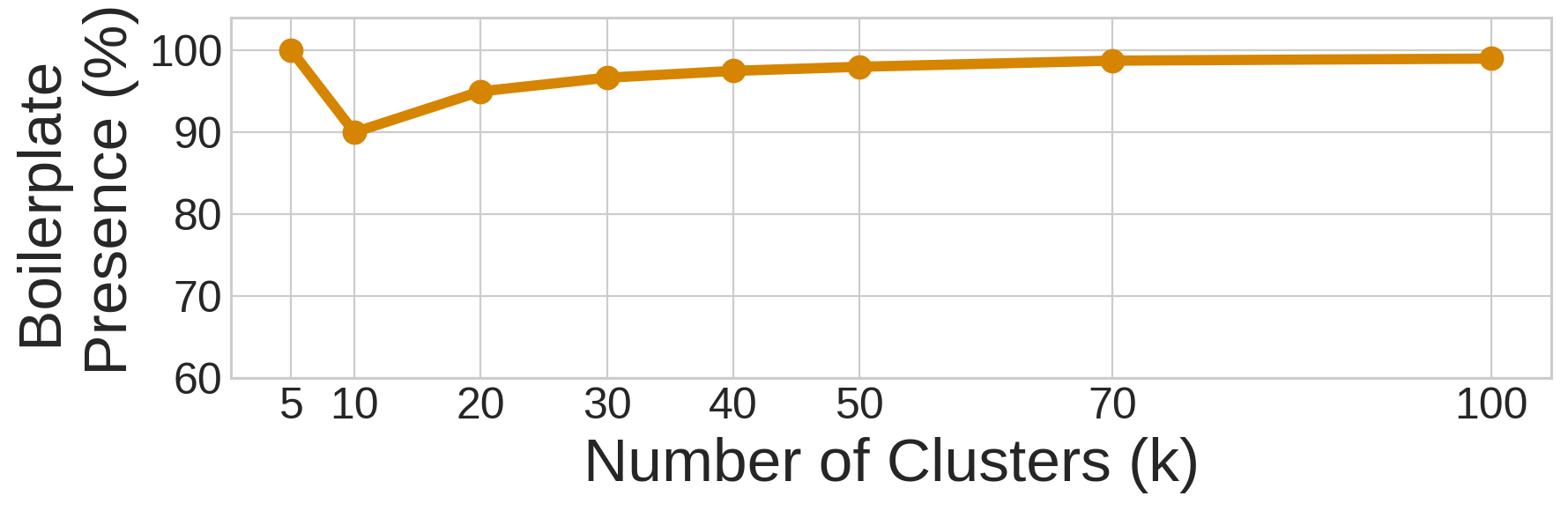}
    \caption{Boilerplate detection rate across different k-means cluster sizes.}
    \label{fig:boiler-kmeans-results}
\end{figure}

To further investigate this phenomenon, we applied k-means clustering to group documents based on their embedding similarity. For each cluster, we extracted the centroid document and its top-$N$ nearest neighbors, and provided them to our algorithm (excluding the final boilerplate extraction step)\footnote{We used the gpt-5-mini-2025-08-07 model.} to determine whether the central document appeared to be generated from a boilerplate. The resulting predictions were manually inspected in order to verify true boilerplate presence and to discard false positives. We repeated the process for $k \in \{5, 10, 20, 30, 40, 50, 70, 100\}$, allowing us to assess how boilerplate detection behaves under varying levels of granularity. The proportion of clusters whose centroids were confirmed to contain boilerplate text is presented in Figure \ref{fig:boiler-kmeans-results}.

From Figure \ref{fig:boiler-kmeans-results} we observe that the vast majority of k-means centroids systematically contain boilerplate. As k increases, we consistently uncover more boilerplate templates across clusters, with an ever larger share of centroids being classified as boilerplate-based. This indicates that increasing clustering granularity reveals additional templated structures rather than reducing their prevalence.



\section{Quantitative Evaluation: Citizen-Oriented Question Answering over \textit{Diavgeia} via Retrieval Augmented Generation}

To demonstrate the practical utility of the corpus for transparency and civic access, we formulate a retrieval-augmented generation (RAG) task in the form of a citizen-facing question answering system over \textit{Diavgeia}. 
The goal is to answer natural-language questions about Greek government decisions, while grounding responses in the underlying acts and explicitly citing their unique identifiers (ADA codes).

\subsection{Task Definition}

Given a user query $q$ in Greek (e.g., "Πόσα χρήματα διατέθηκαν για συμβάσεις τον Ιανουάριο 2024;"), the system must:
(i) retrieve a small set of relevant decisions from the \textit{Diavgeia} corpus,
(ii) construct an evidence context from their text and metadata, and
(iii) generate an answer $a$ that is both correct and explicitly supported by the retrieved decisions, including ADA citations.
The task assumes no prior legal expertise on the user side and is designed to support information needs such as amounts spent, responsible agencies, dates of publication, and descriptions of specific administrative acts.

\subsection{System Implementation}

The deployed system follows a standard RAG architecture \cite{lewis2020rag} adapted to the \textit{Diavgeia} setting. All decisions collected via the official OpenData API are converted into structured text files that concatenate normalized metadata (e.g., ADA, protocol number, issue date, subject, organization, signers, financial fields) with the full PDF-extracted content. These documents are indexed in an Elasticsearch 7.11 instance using the Greek language analyzer, with BM25 as the ranking function.

At query time, we construct a retrieval query by concatenating the current user question with up to five previous turns from the ongoing conversation, enabling basic follow-up and context-aware queries. Elasticsearch performs a full-text BM25 search over the ``content'' field of the index and returns the top-$k$ documents (we use $k=8$ in our prototype). For each hit, we keep both the ADA identifier and the corresponding text segment, and we assemble a compact evidence context by concatenating the metadata header and a subset of the document text.

The generation component uses OpenAI GPT-4o-mini as the primary model, with GPT-4o employed for experiments requiring structured outputs. The prompt consists of: (i) a system message instructing the model to act as an assistant for \textit{Diavgeia} documents and to always reference ADA codes, (ii) a short conversation history, (iii) the retrieved evidence, and (iv) the current user question. The model then produces either a streaming answer (for the interactive chat interface) or a parsed response conforming to a Pydantic schema with both a concise answer and a more detailed explanation including citations. All user–assistant exchanges, along with the detailed responses, are stored in Redis as compressed JSON, enabling stateful dialogue across turns.

\subsection{Evaluation: Automated Question-Answer Pair Creation}
To assess the accuracy and reliability of the RAG system, a comprehensive evaluation framework was implemented using a combination of automated question generation and semantic comparison techniques.
The evaluation dataset was generated using GPT-4.1-mini to create domain-specific question-answer pairs from actual government documents. 
This approach ensures: i) Questions are grounded in real document content, ii) Ground truth answers are accurate and verifiable, iii) Questions cover diverse aspects (decisions, actions, participants, financial data, organizational information).
We select 500 documents from the indexed \textit{Diavgeia} corpus. 
For each document, the LLM generates:
i) One specific, detailed question based on document content, ii) Ground truth answer with relevant details and iii) ADA reference number for traceability.
Accuracy of 66.6\% indicates the RAG system successfully retrieves and generates correct information for approximately two-thirds of queries. 
We also observe strong performance on financial queries (67\% amount match), while moderate semantic similarity suggests responses capture key information even when phrasing differs.
Summary statistics are shown in Table \ref{tab:summary-stats}.

\begin{table}[h!]
\centering
\caption{Summary Statistics for \textit{Diavgeia} QA Dataset (500 Question--Answer Pairs)}
\label{tab:summary-stats}
\begin{tabular}{l r}
\hline
\textbf{Metric} & \textbf{Value} \\
\hline
Total Comparisons & 500 \\
Semantically Equivalent ($\geq 70\%$) & 333 (66.6\%) \\
Not Equivalent ($< 70\%$) & 167 (33.4\%) \\
\hline
Average Semantic Score & 63.0\% \\
Average TF-IDF Similarity & 54.2\% \\
Average Amount Match & 67.0\% \\
\hline
\end{tabular}
\end{table}

\subsection{Manual Evaluation: Multi-Document Aggregation Queries}
\label{subsec:manual-eval}

To complement the automated evaluation and to specifically assess the system's performance on complex multi-document queries, we conducted a manual evaluation focusing on organizational-level questions that require aggregation across multiple decisions. This setting reflects realistic user needs where information about a public entity is scattered across numerous \textit{Diavgeia} documents.

\paragraph{Methodology.}

The scenarios simulate real-world use cases in which users request organization-wide insights. Such queries require retrieving multiple related decisions, synthesizing their content, and performing higher-level aggregation. This simultaneously tests retrieval breadth, cross-document consistency, amount aggregation, signer extraction, and topic summarization.

\textbf{Organization Selection.}
Five diverse government entities were selected to ensure variation in document types, amounts, and administrative roles:
\begin{itemize}
    \item Κρατική Σχολή Ορχηστικής Τέχνης (5 decisions),
    \item Δ.Ε.Υ.Α Θήρας (5 decisions),
    \item Ερευνητικό Κέντρο Καινοτομίας [...] «Αθηνά» (6 decisions),
    \item Περιφερειακό Γενικό Νοσοκομείο Αθηνών (Λαϊκό) (3 decisions),
    \item Εγνατία Οδός (1 decision).
\end{itemize}

\textbf{Question Template.}
Each organization was evaluated using four standardized queries:
\begin{enumerate}
    \item ``How many and which decisions relate to organization X?''
    \item ``What is the total amount spent for these decisions?''
    \item ``Who are the signers for decisions related to the organization?''
    \item ``What are the subjects/topics of these decisions?''
\end{enumerate}

\textbf{Ground Truth Extraction.}
Ground truth was manually derived from the indexed documents. This included:
\begin{itemize}
    \item counting and listing all ADA codes for each organization,
    \item computing the total sum of financial amounts,
    \item compiling a list of unique signers and their positions,
    \item summarizing the topics of the corresponding decisions.
\end{itemize}

\textbf{Response Collection.}
All queries were submitted through the web application interface. Responses were recorded in real time, and conversation context was intentionally preserved across related questions to reflect typical user interaction flows.

\textbf{Human Evaluation.}
Three dimensions were assessed manually: completeness, accuracy, and reasoning quality. Each response was compared directly against the ground truth, accompanied by a qualitative analysis to identify strengths (e.g., correct multi-document synthesis) and weaknesses (e.g., missed decisions, arithmetic errors, partial signer extraction). The overall statistics are shown in Table \ref{tab:manual_eval_overall} and a full example is shown in Table \ref{tab:org1_ksot}.

\begin{table}[ht]
\centering
\caption{Stats on the Manual Multi-Document Evaluation}
\label{tab:manual_eval_overall}
\begin{tabular}{l r l}
\hline
\textbf{Metric} & \textbf{Result} & \textbf{Notes} \\
\hline
Total Questions        & 20       & 4 questions $\times$ 5 organizations \\
Fully Correct          & 14       & 70.0\% \\
Partially Correct      & 3        & 15.0\% \\
Incorrect              & 3        & 15.0\% \\
Overall Accuracy       & 85.0\%   & (Fully + 0.5$\times$Partially) \\
\hline
\end{tabular}
\end{table}

\subsection{Summary}
The RAG system demonstrates strong performance across both evaluation paradigms. In the automated single-document evaluation, it achieves 66.6\% accuracy on 500 diverse queries, with particularly strong results in financial extraction and reliable document identification through ADA codes. 
The manual multi-document evaluation further highlights the system’s capabilities, reaching 85.0\% accuracy on 20 organizational queries, with perfect signer identification (100\%) and perfect topic summarization (100\%), though arithmetic aggregation remains a weaker point (60\%). 
Overall, the combined assessment shows that the system excels at retrieval and synthesis across heterogeneous administrative documents, while numerical reasoning and cross-document arithmetic remain areas for improvement.

\begin{table*}[ht]
\centering
\caption{Results by Organization -- ΚΡΑΤΙΚΗ ΣΧΟΛΗ ΟΡΧΗΣΤΙΚΗΣ ΤΕΧΝΗΣ (5 decisions)}
\label{tab:org1_ksot}
\begin{tabularx}{\textwidth}{lXXl}
\hline
\textbf{Question} & \textbf{Ground Truth} & \textbf{Chatbot Response} & \textbf{Assessment} \\
\hline
Count \& List &
5 decisions: ΡΦ9Υ469ΗΥΖ-6ΩΛ; ΕΖΖΠ469ΗΥΖ-ΧΩ5; 6Ε6Ι469ΗΥΖ-1ΕΘ; ΨΒ41469ΗΥΖ-7ΝΣ; 9ΥΛ9469ΗΥΖ-Ι19 &
5 decisions: ΕΖΖΠ469ΗΥΖ-ΧΩ5, 9ΥΛ9469ΗΥΖ-Ι19, 6Ε6Ι469ΗΥΖ-1ΕΘ, ΨΒ41469ΗΥΖ-7ΝΣ, ΡΦ9Υ469ΗΥΖ-6ΩΛ &
Perfect match -- All 5 ADAs identified \\
Total Amount &
73.225,56 € &
52.736,56 + 10.416,00 + 60,00 + 8.680,00 + 1.333,00 = 73.225,56 € &
Excellent -- Correct sum with breakdown \\
Signers &
Κόκκινος Χαράλαμπος, Διευθυντής &
Χαράλαμπος Κόκκινος, Διευθυντής &
Correct -- Name order variation acceptable \\
Topics &
(Various decision topics) &
Rental, library services, software maintenance, legal services, past dues &
Accurate summary \\
\hline
\multicolumn{4}{l}{\textbf{Performance:} 4/4 correct -- Exemplary multi-document handling} \\
\hline
\end{tabularx}
\end{table*}



\subsection{Scope and Limitations}

The current implementation indexes a subset of \textit{Diavgeia} decisions and uses purely lexical BM25 retrieval, which may miss semantically relevant acts when queries are vague or paraphrased. The generator is constrained by a 1,500-token output limit and can occasionally over-generalise when evidence is sparse or noisy. Nonetheless, the system already supports a broad range of real-world queries (financial, organizational, temporal, and document-oriented) and serves as a reference implementation of a \textit{Diavgeia} RAG task that future work can extend with hybrid retrieval, domain-adapted Greek language models, and richer evaluation protocols.

\section{Challenges and Future Directions}

Although our work lays the groundwork for computational transparency, there are still a number of obstacles to overcome.
Robust handling of noisy OCR, varied document formats, and changing administrative schemas is necessary to guarantee the dependability of AI-driven studies. 
Additionally, models must be built to reduce hallucinations, maintain legal accuracy, and offer traceable reasoning—particularly when applied in high-stakes public sector situations. 
New techniques for verified reasoning, governance-aware language models, and controlled multi-agent interaction will be necessary to move towards novel AI systems.
Future studies should investigate human-in-the-loop verification, counterfactual simulation, and multi-perspective oversight systems, as well as how these techniques might be included into formal auditing processes and cross-national transparency databases.  
When taken as a whole, these paths show the way towards reliable, scalable AI technologies that significantly improve transparency and accountability in contemporary governance.
Next, we present two such systems that show exceptional potential for advancing transparency.

\subsection{AI-Driven Public Decision Auditor}

A promising and fundamentally new research direction enabled by large-scale open-government corpora such as \textit{Diavgeia} is the development of an \textbf{AI Public Decision Auditor}: an automated system that performs continuous, data-driven oversight of government actions in real time. Rather than supporting only retrospective analysis or simple information access, such a system would operationalise proactive transparency by detecting irregularities, inconsistencies, and atypical behaviour at the moment decisions are published.

Recent work on knowledge-graph-based procurement platforms and anomaly detection has shown that open data can be used to identify unusual tenders, suspicious supplier patterns, and potential integrity risks in public spending \cite{soylu2020enhancing, soylu2021theybuyforyou}. 
Machine learning has been applied to open-government data to detect improprieties and estimate corruption risk in public resource allocation \citep{vaqueiro2023machine}. 
In parallel, policy and governance communities have started to articulate frameworks for algorithmic transparency and accountability in the public sector \citep{oecd2024_algo_transparency,ainow2021_public_sector_accountability}, and to explore how AI reshapes public auditing and oversight institutions \citep{genaro_moya2025_ai_auditing}.

An AI Public Decision Auditor would build on these foundations while going beyond domain-specific anomaly detection toward holistic, cross-cutting monitoring of the full decision stream. It would leverage statistical modelling, temporal pattern analysis, and large language models to identify signals of administrative risk or unusual institutional behaviour. Core analytical capabilities could include: \textbf{i) Anomaly detection in spending and procurement}: automatically flagging unusual financial patterns, inconsistent justifications, or deviations from agency- and category-specific historical norms, extending existing procurement-oriented anomaly detectors to the broader space of administrative acts. \textbf{ii) Inconsistency tracing across time}: detecting when newly published decisions contradict past rulings, commitments, or procedural habits of the same public body, or depart from legally relevant precedents.
\textbf{iii) Procedural integrity checks}: identifying potentially irregular approvals, including missing or anomalous signatures, atypical approval chains, rushed publication timelines, or mismatches between an act's declared type and its legal or substantive content.
\textbf{iv) Temporal and political clustering}: highlighting clusters of decisions that correlate suspiciously with election cycles, budget deadlines, procurement periods, or other political and administrative events, thereby supporting risk-based oversight of politically sensitive windows.

By transforming continuous streams of public-sector decisions into actionable, explainable signals, an AI Public Decision Auditor would establish a machine-driven accountability layer that is accessible to citizens, journalists, researchers, and watchdog organisations—even without technical expertise. This vision moves beyond search, retrieval, or question answering, positioning open-government data as the substrate for \emph{real-time, anticipatory oversight}. As such, it represents a radical and societally impactful next step for transparency and AI-for-governance research.

\subsection{LLM-Based Multi-Agent Collective Oversight Simulator}

Beyond single-agent auditors, a radical next step is a LLM-based multi-agent Collective Oversight Simulator for public decisions. In this paradigm, newly published administrative acts (e.g., \textit{Diavgeia} decisions) are injected into a simulated ecosystem of specialized LLM agents. 
Each agent embodies a distinct institutional role or normative objective, and the agents \emph{debate}, \emph{negotiate}, and \emph{stress-test} the decision before it is surfaced to end-users. Rather than passively flagging anomalies, the system performs proactive, scenario-based oversight, exposing weaknesses, and inconsistencies that may only emerge through interaction between multiple perspectives.

This vision builds on recent work showing that LLM-based multi-agent systems can autonomously cooperate, role-play, and coordinate on complex tasks \citep{wu2023autogen, guo2024large}.
Role-playing frameworks such as CAMEL \cite{li2023camel} demonstrate how distinct agent profiles and inception prompts can elicit complementary behaviours from the same underlying model, while general multi-agent conversation frameworks like AutoGen enable programmable interaction patterns between specialised agents and tools \citep{wu2023autogen}. 
Multi-agent debate architectures further suggest that structured argumentation between LLMs can improve reasoning quality and truth-seeking \citep{hu2025debate}, and recent surveys highlight the particular relevance of LLM-based multi-agent systems to domains requiring diverse expertise, such as legal and public-sector decision-making \citep{jiang2024agents, chen2025debate}.

Within this landscape, the Collective Oversight Simulator instantiates a set of persistent, LLM-based agents, for example:
\textbf{Legal Coherence Agent}: checks compatibility with existing legislation, prior decisions, and cross-references, proposing counterfactuals where minor changes to the act would trigger conflicts.
\textbf{i) Budget Compliance Agent}: simulates fiscal impact over time, tests alternative budgetary scenarios, and highlights combinations of decisions that risk overruns or hidden commitments.
\textbf{ii) Procedural Integrity Agent}: systematically perturbs timelines, signatory chains, and act types to identify minimal changes that render the act procedurally irregular, thereby mapping its ``fragility'' with respect to administrative rules.
\textbf{iii) Citizen Fairness Agent}: reasons about distributional effects across population groups, using fairness-inspired heuristics and counterfactual cases to surface potential inequities or disparate impacts.

Agents engage in a structured dialogue procedure in which they challenge one another's presumptions, suggest counterfactual changes to the choice, and either converge or fail to converge on a set of risk statements and justifications.  
The result is a multi-perspective report that identifies areas of agreement and disagreement among institutional responsibilities rather than a single score.  
In terms of methodology, this creates a new field of work at the nexus of computational public administration and LLM-based multi-agent systems: creating role specifications, debate protocols, and evaluation frameworks that make simulated oversight useful for practitioners and auditable by citizens, journalists, and regulators.

\section{Conclusion}

This work presented the first large-scale, processed corpus of Greek government decisions and demonstrated its value for computational transparency research. Our contributions include: (i) \textbf{a one-million–document open dataset} with normalized metadata, canonical identifiers, and high-quality Markdown text extracted from PDFs and XML; (ii) \textbf{a fully reproducible extraction pipeline} for harvesting, cleaning, and structuring \textit{Diavgeia} documents at Web scale; (iii) \textbf{qualitative analyses} via boilerplate detection; and (iv) \textbf{a retrieval-augmented generation (RAG) benchmark} with representative questions, curated answers, and baseline performance, simulating \textbf{a citizen-facing tool} enabling natural-language querying of government decisions, demonstrating the dataset’s practical value for transparency and civic empowerment.

Beyond these contributions, our findings highlight the transformative potential of AI as an accountability mechanism within digital governance ecosystems. 
As language models grow more capable of reasoning over long documents, cross-referencing administrative histories, and detecting irregularities, they can function as computational aides that continuously surface inconsistencies, atypical patterns, or procedural anomalies across vast public decision streams. 
The scale and structure of the dataset released here provide a foundation for developing such systems in a rigorous and data-driven manner, especially in the case of low- to mid-resource languages like Greek, where limited NLP data are available \cite{outsios2018wordembeddingslargescalegreek, outsios-etal-2020-evaluation}.

Looking forward, future models may incorporate role-specific reasoning, temporal analytics, counterfactual simulation, and debate-style multi-agent interaction to evaluate decisions from legal, financial, procedural, and fairness perspectives. 
These advances could give rise to proactive oversight tools—LLM-based auditors and multi-agent governance simulators—that support citizens, journalists, regulators, and civil-society organizations in holding institutions accountable. 




\bibliographystyle{ACM-Reference-Format}
\bibliography{bibliography}

@article{serderidis2024d2kg,
  title={d2kg: An integrated ontology for knowledge graph-based representation of government decisions and acts: The Greek Programme Diavgeia case},
  author={Serderidis, Konstantinos and Konstantinidis, Ioannis and Meditskos, Georgios and Peristeras, Vassilios and Bassiliades, Nick},
  journal={Semantic Web},
  volume={15},
  number={5},
  pages={1677--1699},
  year={2024},
  publisher={SAGE Publications Sage UK: London, England}
}

@article{vafopoulos2012publicspending,
  title={Public spending: Interconnecting and visualizing greek public expenditure following linked open data directives},
  author={Vafopoulos, Michalis N and Meimaris, Marios and Papantoniou, Agis and Anagnostopoulos, Ioannis and Alexiou, Giorgos and Avraam, Ioannis and Xidias, Ioannis and Vafeiadis, Giorgos and Loumos, Vasilis},
  journal={Available at SSRN 2064517},
  year={2012}
}

@article{soylu2021theybuyforyou,
  title={TheyBuyForYou platform and knowledge graph: Expanding horizons in public procurement with open linked data},
  author={Soylu, Ahmet and Corcho, Oscar and Elves{\ae}ter, Brian and Badenes-Olmedo, Carlos and Blount, Tom and Yedro Mart{\'\i}nez, Francisco and Kovacic, Matej and Posinkovic, Matej and Makgill, Ian and Taggart, Chris and others},
  journal={Semantic Web},
  volume={13},
  number={2},
  pages={265--291},
  year={2022},
  publisher={SAGE Publications Sage UK: London, England}
}

@inproceedings{sherif2014nif4oggd,
  title={NIF4OGGD-NLP Interchange Format for Open German Governmental Data.},
  author={Sherif, Mohamed Ahmed and Coelho, Sandro Atha{\'\i}de and Usbeck, Ricardo and Hellmann, Sebastian and Lehmann, Jens and Br{\"u}mmer, Martin and Both, Andreas},
  booktitle={LREC},
  pages={3524--3528},
  year={2014}
}

@inproceedings{abrami2024german,
  title={German parliamentary corpus (GerParCor) reloaded},
  author={Abrami, Giuseppe and Bagci, Mevl{\"u}t and Mehler, Alexander},
  booktitle={Proceedings of the 2024 Joint International Conference on Computational Linguistics, Language Resources and Evaluation (LREC-COLING 2024)},
  pages={7707--7716},
  year={2024}
}

@inproceedings{viira2024parlamint,
  title={ParlaMint Widened: a European Dataset of Freedom of Information Act Documents (Position Paper)},
  author={Viira, Gerda and Marx, Maarten and Larooij, Maik},
  booktitle={Proceedings of the IV Workshop on Creating, Analysing, and Increasing Accessibility of Parliamentary Corpora (ParlaCLARIN)@ LREC-COLING 2024},
  pages={171--172},
  year={2024}
}

@article{zuiderwijk2014opendata,
  title={Innovation with open data: Essential elements of open data ecosystems},
  author={Zuiderwijk, Anneke and Janssen, Marijn and Davis, Chris},
  journal={Information polity},
  volume={19},
  number={1-2},
  pages={17--33},
  year={2014},
  publisher={SAGE Publications Sage UK: London, England}
}

@article{bertot2010transparency,
  title={Using ICTs to create a culture of transparency: E-government and social media as openness and anti-corruption tools for societies},
  author={Bertot, John C and Jaeger, Paul T and Grimes, Justin M},
  journal={Government information quarterly},
  volume={27},
  number={3},
  pages={264--271},
  year={2010},
  publisher={Elsevier}
}

@misc{edp2020maturity,
  title={Open Data Maturity Report},
  author={{European Data Portal}},
  year={2020},
  howpublished={European Commission}
}

@inproceedings{chalkidis2019large,
  title={Large-scale multi-label text classification on EU legislation},
  author={Chalkidis, Ilias and Fergadiotis, Emmanouil and Malakasiotis, Prodromos and Androutsopoulos, Ion},
  booktitle={Proceedings of the 57th annual meeting of the association for computational linguistics},
  pages={6314--6322},
  year={2019}
}

@inproceedings{chalkidis2022lexglue,
  title={LexGLUE: A benchmark dataset for legal language understanding in English},
  author={Chalkidis, Ilias and Jana, Abhik and Hartung, Dirk and Bommarito, Michael and Androutsopoulos, Ion and Katz, Daniel and Aletras, Nikolaos},
  booktitle={Proceedings of the 60th Annual Meeting of the Association for Computational Linguistics (Volume 1: Long Papers)},
  pages={4310--4330},
  year={2022}
}

@article{lewis2020rag,
  title={Retrieval-augmented generation for knowledge-intensive nlp tasks},
  author={Lewis, Patrick and Perez, Ethan and Piktus, Aleksandra and Petroni, Fabio and Karpukhin, Vladimir and Goyal, Naman and K{\"u}ttler, Heinrich and Lewis, Mike and Yih, Wen-tau and Rockt{\"a}schel, Tim and others},
  journal={Advances in neural information processing systems},
  volume={33},
  pages={9459--9474},
  year={2020}
}

@inproceedings{soylu2020enhancing,
  title={Enhancing public procurement in the European Union through constructing and exploiting an integrated knowledge graph},
  author={Soylu, Ahmet and Corcho, Oscar and Elves{\ae}ter, Brian and Badenes-Olmedo, Carlos and Mart{\'\i}nez, Francisco Yedro and Kovacic, Matej and Posinkovic, Matej and Makgill, Ian and Taggart, Chris and Simperl, Elena and others},
  booktitle={International Semantic Web Conference},
  pages={430--446},
  year={2020},
  organization={Springer}
}

@inproceedings{vaqueiro2023machine,
  title={Machine Learning Applied to Open Government Data for the Detection of Improprieties in the Application of Public Resources},
  author={Vaqueiro, Ramon and Vargas, Ana and Escovedo, Tatiana and Kalinowski, Marcos},
  booktitle={Proceedings of the XIX Brazilian Symposium on Information Systems},
  pages={213--220},
  year={2023}
}

@techreport{oecd2024_algo_transparency,
  title        = {Algorithmic Transparency in the Public Sector: Emerging Practices and Policy Guidelines},
  author       = {{OECD}},
  institution  = {Organisation for Economic Co-operation and Development},
  year         = {2024},
  url          = {https://oecd.org}
}

@techreport{ainow2021_public_sector_accountability,
  title        = {Government by Algorithm: Public-Sector Algorithmic Accountability},
  author       = {{AI Now Institute}},
  institution  = {AI Now Institute},
  year         = {2021},
  url          = {https://ainowinstitute.org}
}

@Article{genaro_moya2025_ai_auditing,
AUTHOR = {Genaro-Moya, Dolores and López-Hernández, Antonio Manuel and Godz, Mariia},
TITLE = {Artificial Intelligence and Public Sector Auditing: Challenges and Opportunities for Supreme Audit Institutions},
JOURNAL = {World},
VOLUME = {6},
YEAR = {2025},
NUMBER = {2},
ARTICLE-NUMBER = {78},
URL = {https://www.mdpi.com/2673-4060/6/2/78},
ISSN = {2673-4060},
DOI = {10.3390/world6020078}
}

@article{li2023camel,
  title={Camel: Communicative agents for" mind" exploration of large language model society},
  author={Li, Guohao and Hammoud, Hasan and Itani, Hani and Khizbullin, Dmitrii and Ghanem, Bernard},
  journal={Advances in Neural Information Processing Systems},
  volume={36},
  pages={51991--52008},
  year={2023}
}

@inproceedings{wu2023autogen,
  title={Autogen: Enabling next-gen LLM applications via multi-agent conversations},
  author={Wu, Qingyun and Bansal, Gagan and Zhang, Jieyu and Wu, Yiran and Li, Beibin and Zhu, Erkang and Jiang, Li and Zhang, Xiaoyun and Zhang, Shaokun and Liu, Jiale and others},
  booktitle={First Conference on Language Modeling},
  year={2024}
}

@inproceedings{hu2025debate,
  title={Debate-to-write: A persona-driven multi-agent framework for diverse argument generation},
  author={Hu, Zhe and Chan, Hou Pong and Li, Jing and Yin, Yu},
  booktitle={Proceedings of the 31st International Conference on Computational Linguistics},
  pages={4689--4703},
  year={2025}
}

@inproceedings{leonhardt2020boilerplate,
  title={Boilerplate removal using a neural sequence labeling model},
  author={Leonhardt, Jurek and Anand, Avishek and Khosla, Megha},
  booktitle={Companion Proceedings of the Web Conference 2020},
  pages={226--229},
  year={2020}
}

@inproceedings{vogels2018web2text,
  title={Web2text: Deep structured boilerplate removal},
  author={Vogels, Thijs and Ganea, Octavian-Eugen and Eickhoff, Carsten},
  booktitle={European Conference on Information Retrieval},
  pages={167--179},
  year={2018},
  organization={Springer}
}

@inproceedings{beris2018modeling,
  title={Modeling and preserving Greek government decisions using semantic web technologies and permissionless blockchains},
  author={Beris, Themis and Koubarakis, Manolis},
  booktitle={European Semantic Web Conference},
  pages={81--96},
  year={2018},
  organization={Springer}
}

@article{stylianou2022doc2kg,
  title={Doc2KG: Transforming document repositories to knowledge graphs},
  author={Stylianou, Nikolaos and Vlachava, Danai and Konstantinidis, Ioannis and Bassiliades, Nick and Peristeras, Vassilios},
  journal={International Journal on Semantic Web and Information Systems (IJSWIS)},
  volume={18},
  number={1},
  pages={1--20},
  year={2022},
  publisher={IGI Global Scientific Publishing}
}

@article{chen2023bm25,
  title={Bm25 query augmentation learned end-to-end},
  author={Chen, Xiaoyin and Wiseman, Sam},
  journal={arXiv preprint arXiv:2305.14087},
  year={2023}
}

@article{jiang2024agents,
  title={Agents on the Bench: Large Language Model Based Multi Agent Framework for Trustworthy Digital Justice},
  author={Jiang, Cong and Yang, Xiaolei},
  journal={arXiv preprint arXiv:2412.18697},
  year={2024}
}

@article{chen2025debate,
  title={Debate-feedback: A multi-agent framework for efficient legal judgment prediction},
  author={Chen, Xi and Mao, Mao and Li, Shuo and Shangguan, Haotian},
  journal={arXiv preprint arXiv:2504.05358},
  year={2025}
}

@article{guo2024large,
  title={Large language model based multi-agents: A survey of progress and challenges},
  author={Guo, Taicheng and Chen, Xiuying and Wang, Yaqi and Chang, Ruidi and Pei, Shichao and Chawla, Nitesh V and Wiest, Olaf and Zhang, Xiangliang},
  journal={arXiv preprint arXiv:2402.01680},
  year={2024}
}

@inproceedings{outsios-etal-2020-evaluation,
    title = "Evaluation of {G}reek Word Embeddings",
    author = "Outsios, Stamatis  and
      Karatsalos, Christos  and
      Skianis, Konstantinos  and
      Vazirgiannis, Michalis",
    booktitle = "Proceedings of the Twelfth Language Resources and Evaluation Conference",
    month = may,
    year = "2020",
    address = "Marseille, France",
    publisher = "European Language Resources Association",
    url = "https://aclanthology.org/2020.lrec-1.310/",
    pages = "2543--2551",
    language = "eng",
    ISBN = "979-10-95546-34-4",
}

@misc{outsios2018wordembeddingslargescalegreek,
      title={Word Embeddings from Large-Scale Greek Web Content}, 
      author={Stamatis Outsios and Konstantinos Skianis and Polykarpos Meladianos and Christos Xypolopoulos and Michalis Vazirgiannis},
      year={2018},
      eprint={1810.06694},
      archivePrefix={arXiv},
      primaryClass={cs.CL},
      url={https://arxiv.org/abs/1810.06694}, 
}

\appendix

\section{Appendix}

Next, we present some additional description of fields, analysis, and results.

Table \ref{tab:diavgeia-data} summarizes the core metadata fields available in \textit{Diavgeia} decision records. For each field, we include the original Greek description as provided by the platform, along with an English translation to support international accessibility and reproducibility. These fields capture identifiers, organizational information, timestamps, status indicators, and additional type-specific metadata, forming the basis for structured analysis of government decisions.

\begin{table*}[h]
\centering
\caption{Description of Decision Fields}
\label{tab:diavgeia-data}
\resizebox{\textwidth}{!}{%
\begin{tabular}{|l|l|l|}
\hline
\textbf{Field} & \textbf{Description (GR)} & \textbf{English Translation} \\ \hline
ada & Αριθμός Διαδικτυακής Ανάρτησης & Web Posting Number \\ \hline
protocolNumber & Αριθμός Πρωτοκόλλου & Protocol Number \\ \hline
subject & Θέμα πράξης & Subject of the decision \\ \hline
Issue Date & Ημερομηνία έκδοσης σε μορφή Unix Timestamp & Issue date (Unix timestamp, ms) \\ \hline
decisionTypeId & Κωδικός του τύπου πράξης & Decision type ID \\ \hline
organizationId & Κωδικός φορέα & Organization ID \\ \hline
unitIds & Λίστα κωδικών των μονάδων που εμπλέκονται & IDs of units involved \\ \hline
signerIds & Λίστα κωδικών των τελικών υπογραφόντων & IDs of official signatories \\ \hline
extraFieldValues & Ειδικά πεδία ανάλογα με τον τύπο πράξης & Type-specific metadata fields \\ \hline
submissionTimestamp & Ημερομηνία/ώρα τελευταίας τροποποίησης σε Unix Timestamp & Last modification timestamp (ms) \\ \hline
status & Κατάσταση πράξης (PUBLISHED, PENDING REVOCATION, REVOKED, SUBMITTED etc.) & Status of the decision \\ \hline
versionId & Αριθμός έκδοσης & Version number \\ \hline
\end{tabular}
}
\end{table*}

Table \ref{tab:ocr_speed_comparison} presents a comparative overview of the execution time of different PDF text-extraction and OCR methods. It highlights the substantial performance differences between native PDF extraction tools, lightweight OCR engines, and more computationally intensive deep-learning-based OCR pipelines. By measuring the total processing time under identical conditions, the table  helps identify which methods are most suitable for large-scale processing of Diavgeia documents.

\begin{table*}[h]
\centering
\caption{Speed Comparison of PDF Extraction and OCR Methods}
\label{tab:ocr_speed_comparison}
\begin{tabular}{l r r r}
\hline
\textbf{Method} & \textbf{Total Time (s)} & \textbf{Avg.\ Time per Page (s)} & \textbf{Relative Speed} \\
\hline
Extraction (PyMuPDF)        & 0.07      & $\sim$0.00017 & $\sim$2500$\times$ Faster \\
Paddle OCR (Zoom~2)         & 423.69    & 0.97          & 1$\times$ \\
Tesseract OCR (Zoom~2)      & 429.43    & 0.99          & 0.97$\times$ \\
Tesseract OCR (Zoom~3)      & 523.76    & 1.20          & 0.81$\times$ \\
Paddle OCR (Zoom~3)         & 896.10    & 2.06          & 0.47$\times$ \\
GPT-5-nano (Ground Truth)   & $\sim$25{,}200 (7 hrs) & $\sim$126 & 0.007$\times$ \\
\hline
\end{tabular}
\end{table*}

Tables \ref{tab:org2_deya_thiras}, \ref{tab:org3_athena}, \ref{tab:org4_laiko} show results by organization on manual evaluation regarding multi-document aggregation queries.
Our evaluation highlights clear differences in performance across tasks. 
Signer identification achieves perfect accuracy, consistently extracting names and positions across organizations and handling name variations without errors. Topic summarization also performs exceptionally well, producing coherent and semantically accurate summaries, sometimes even offering more detail than the reference summaries. 
Document enumeration performs well overall, correctly identifying most relevant decisions, though it occasionally retrieves extra or incorrect documents—an issue that appears mainly when organization names are not unique. 
Financial aggregation is the weakest area, succeeding only when monetary amounts are explicitly listed. It struggles with implicit calculations or incomplete data, and a large deviation in the ΛΑΙΚΟ case (28\% error) suggests a systematic limitation in numerically grounded reasoning across financial documents.

\begin{table*}[h]
\centering
\caption{Results by Organization -- Δ.Ε.Υ.Α ΘΗΡΑΣ (5 decisions)}
\label{tab:org2_deya_thiras}
\begin{tabularx}{\textwidth}{lXXl}
\hline
\textbf{Question} & \textbf{Ground Truth} & \textbf{Chatbot Response} & \textbf{Assessment} \\
\hline
Count \& List &
5 decisions: ΨΣ02ΟΕΨΠ-ΛΔΤ; 6Μ6ΨΟΕΨΠ-ΛΗ2; 98Π9ΟΕΨΠ-ΞΨΕ; 66ΤΙΟΕΨΠ-ΕΟΤ; 6Λ98ΟΕΨΠ-ΩΣΙ &
6 decisions: ΨΣ02ΟΕΨΠ-ΛΔΤ, 6Μ6ΨΟΕΨΠ-ΛΗ2, 98Π9ΟΕΨΠ-ΞΨΕ, 66ΤΙΟΕΨΠ-ΕΟΤ, 6Λ98ΟΕΨΠ-ΩΣΙ + 1 extra &
Minor error -- Retrieved 5 correct + 1 incorrect \\
Total Amount &
381,22 € &
381,22 € &
Perfect \\
Signers &
ΖΩΡΖΟΣ ΕΥΑΓΓΕΛΟΣ, Διευθυντής &
Ευάγγελος Ζώρζος, Προϊστάμενος Τμήματος και Διευθυντής &
Correct with additional detail \\
Topics &
ΚΡΑΤΗΣΕΙΣ ΤΡΑΠΕΖΩΝ &
Κρατήσεις τραπεζών &
Correct \\
\hline
\multicolumn{4}{l}{\textbf{Performance:} 3.5/4 -- One extra document retrieved but calculations correct} \\
\hline
\end{tabularx}
\end{table*}

\begin{table*}[h]
\centering
\caption{Results by Organization -- ΕΡΕΥΝΗΤΙΚΟ ΚΕΝΤΡΟ ΚΑΙΝΟΤΟΜΙΑΣ [...] «ΑΘΗΝΑ» (6 decisions)}
\label{tab:org3_athena}
\begin{tabularx}{\textwidth}{lXXl}
\hline
\textbf{Question} & \textbf{Ground Truth} & \textbf{Chatbot Response} & \textbf{Assessment} \\
\hline
Count \& List &
6 decisions: 6ΥΒΘ469ΗΞΩ-ΣΚΟ; 6ΨΨΞ469ΗΞΩ-ΦΧΗ; 6ΑΝ7469ΗΞΩ-7Ε3; 9Ο5Β469ΗΞΩ-ΦΜΙ; 9ΗΓΕ469ΗΞΩ-0Τ3; ΨΨΝΛ469ΗΞΩ-5ΩΞ &
6 decisions: 6ΨΨΞ469ΗΞΩ-ΦΧΗ, 9Ο5Β469ΗΞΩ-ΦΜΙ, 6ΑΝ7469ΗΞΩ-7Ε3, ΨΨΝΛ469ΗΞΩ-5ΩΞ, 6ΥΒΘ469ΗΞΩ-ΣΚΟ, 6ΥΜΝ469ΗΡΙ-Ρ66 &
Partial -- 5/6 correct, 1 wrong substitution \\
Total Amount &
11.170,33 € &
11.170,33 € &
Perfect despite wrong ADA \\
Signers &
Κατσαμάνης Νάσος, ΙΩΑΝΝΗΣ ΕΜΙΡΗΣ, ΓΑΡΟΦΑΛΑΚΗΣ ΜΙΝΩΣ, Στύλιος Χρυσόστομος &
Νάσος Κατσαμάνης (Αναπληρωτής Διευθυντής), Μίνως Γαροφαλάκης (Διευθυντής ΙΠΣΥ), Ιωάννης Εμίρης (Γενικός Διευθυντής) &
$\sim$ Incomplete -- Missing 1 signer \\
Topics &
(List of 6 topics) &
Software subscriptions, elevator maintenance, H/M installations, cybersecurity subscription, courier services, call for proposals &
Good summary \\
\hline
\multicolumn{4}{l}{\textbf{Performance:} 2.5/4 -- Retrieval issue but good synthesis} \\
\hline
\end{tabularx}
\end{table*}

\begin{table*}[h]
\centering
\caption{Results by Organization -- ΠΕΡ.ΓΕΝ. ΝΟΣΟΚΟΜΕΙΟ ΑΘΗΝΩΝ (ΛΑΙΚΟ) (3 decisions)}
\label{tab:org4_laiko}
\begin{tabularx}{\textwidth}{lXXl}
\hline
\textbf{Question} & \textbf{Ground Truth} & \textbf{Chatbot Response} & \textbf{Assessment} \\
\hline
Count \& List &
3 decisions: 9ΓΨΒ4690Ω8-ΥΤΧ; 6ΝΞΕ4690Ω8-ΑΗΖ; 935Δ4690Ω8-ΑΤΓ &
3 decisions: 6ΝΞΕ4690Ω8-ΑΗΖ, 9ΓΨΒ4690Ω8-ΥΤΧ, 935Δ4690Ω8-ΑΤΓ &
Perfect \\
Total Amount &
338.580,80 € &
243.380,00 € &
Incorrect -- Major calculation error (72\% accuracy) \\
Signers &
ΡΟΪΔΗΣ ΘΕΟΦΑΝΗΣ, Διοικητής &
ΘΕΟΦΑΝΗΣ ΡΟΪΔΗΣ, Διοικητής &
Correct \\
Topics &
ΑΠΟΦΑΣΗ ΑΝΑΛΗΨΗΣ ΠΙΣΤΩΣΗΣ &
Payroll commitments, allowances, ΕΦΚΑ contributions, training, facilities maintenance &
More detailed than GT \\
\hline
\multicolumn{4}{l}{\textbf{Performance:} 3/4 -- Document retrieval perfect but aggregation failed} \\
\hline
\end{tabularx}
\end{table*}

\end{document}